\PassOptionsToPackage{table,xcdraw}{xcolor}

\documentclass[sigconf,9pt]{acmart}

\usepackage{tabularray}
\usepackage[english]{babel}
\usepackage{mathtools}
\usepackage{algorithmic}
\usepackage{graphicx}
\usepackage{textcomp}
\usepackage{siunitx}
\usepackage[acronym]{glossaries}
\usepackage[table]{xcolor}
\usepackage[capitalize, noabbrev]{cleveref}
\usepackage[american, RPvoltages]{circuitikz}
\usepackage{listings}
\usepackage{xspace}
\usepackage{dcolumn, booktabs}
\newcolumntype{d}[1]{D{.}{.}{#1}}
\usepackage{multirow}
\usepackage{tabularray}
\usepackage{rotating}
\usepackage{comment}

\usepackage{enumitem}
\setlist[itemize]{leftmargin=*}

\definecolor{red}    {HTML}{b7211f}
\definecolor{orange} {HTML}{FFA500}
\definecolor{blue}   {HTML}{05568c}
\definecolor{green}  {HTML}{147546}
\definecolor{plum}   {HTML}{DDA0DD}
\definecolor{purple} {HTML}{9370DB}

\graphicspath{{./figures/}{./figures/support/}}
\def\BibTeX{{\rm B\kern-.05em{\sc i\kern-.025em b}\kern-.08em%
    T\kern-.1667em\lower.7ex\hbox{E}\kern-.125emX}}
\usetikzlibrary{fit}
\usetikzlibrary{calc}
\usetikzlibrary{patterns}
\usetikzlibrary{arrows}
\usetikzlibrary{arrows.meta}
\usetikzlibrary{shapes.arrows}
\usetikzlibrary{positioning}
\usetikzlibrary{bending}
\usetikzlibrary{angles}
\usetikzlibrary{quotes}
\usetikzlibrary{decorations.pathmorphing}
\usetikzlibrary{decorations.markings}
\makeatletter
\newcommand\notsotiny{\@setfontsize\notsotiny\@vipt\@viipt}
\makeatother
\makeatletter
\newcommand{\ie}{\emph{i.e.}}

\newcommand*{\wrt}{\emph{w.r.t.}\@\xspace}

\makeatother

\makeatletter
\tikzset{%
    >={Latex[round]},
    every node/.append style={align=center, font=\sffamily\footnotesize},
    AN/.style={anchor=north},
    ANW/.style={anchor=north west},
    ANE/.style={anchor=north east},
    AE/.style={anchor=east},
    AW/.style={anchor=west},
    AS/.style={anchor=south},
    ASW/.style={anchor=south west},
    AC/.style={anchor=center},
    ultra thin/.style= {line width=0.2pt},
    very thin/.style=  {line width=0.4pt},
    thin/.style=       {line width=0.6pt},
    semithick/.style=  {line width=0.8pt},
    thick/.style=      {line width=1.0pt},
    very thick/.style= {line width=1.4pt},
    ultra thick/.style={line width=1.8pt}
    dflt_cylinder_blue/.style={
        cylinder, thick,
        draw=blue!80!black,
        minimum height=3.00cm, minimum width=0.75cm,
        shape border rotate=180,
        cylinder uses custom fill,
        cylinder body fill=blue!10,
        cylinder end  fill=blue!5
    },
    dflt_cylinder_orange/.style={
        cylinder, thick,
        draw=orange!80!black,
        minimum height=3.00cm, minimum width=0.75cm,
        shape border rotate=180,
        cylinder uses custom fill,
        cylinder body fill=orange!10,
        cylinder end  fill=orange!5
    },
    dflt_cylinder_green/.style={
        cylinder, thick,
        draw=green!80!black,
        minimum height=3.00cm, minimum width=0.75cm,
        shape border rotate=180,
        cylinder uses custom fill,
        cylinder body fill=green!10,
        cylinder end  fill=green!5
    },
    draw_red/.style={
        draw=red!80!black, fill=red!10, thick, outer sep=0pt, inner sep=0pt
    },
    draw_blue/.style={
        draw=blue!80!black, fill=blue!10, thick, outer sep=0pt, inner sep=0pt
    },
    draw_orange/.style={
        draw=orange!80!black, fill=orange!10, thick, outer sep=0pt, inner sep=0pt
    },
    draw_green/.style={
        draw=green!80!black, fill=green!10, thick, outer sep=0pt, inner sep=0pt
    },
    draw_yellow/.style={
        draw=yellow!80!black, fill=yellow!10, thick, outer sep=0pt, inner sep=0pt
    },
	up_arrow/.style={
		draw, single arrow,
		rounded corners = 0.5pt,
		minimum width  = 1.00cm,
		minimum height = 0.75cm,
		single arrow head extend = 1mm,
		single arrow tip angle   = 100,
		shape border rotate      = 100
	},
}
\makeatother

\colorlet{number}{black}
\colorlet{keyword}{purple!80!black}
\colorlet{constant}{red}
\colorlet{comment}{green!80!black}
\colorlet{string}{green!60!black}
\colorlet{discipline}{orange!60!gray}

\lstdefinestyle{SPICE}{
    basicstyle          = {\scriptsize\ttfamily},
    belowskip           = -1.5\baselineskip,
    columns             = fixed,
    captionpos          = b,
	numbers				= left,
	numberstyle			= \color{number},
	numbersep			= 5pt,
	frame				= tb,
	xleftmargin			= 1.5em,
	framexleftmargin	= 1.5em,
	tabsize				= 2,
	showtabs			= false,
	showspaces			= false,
	showstringspaces	= false,
	commentstyle		= \color{comment},
	keywordstyle		= \color{keyword}\bfseries,
	stringstyle			= \color{string},
	morekeywords={
		.op,
		.dc,
		.ac,
		.tran,
		.ic,
		.if,
		.nodeset,
		.temp,
		.param,
		.step,
		.alter,
		.print,
		.plot,
		.option,
		.options,
		.measure,
		.meas,
		.model,
		.include,
		.end,
		.subckt,
		.ends,
		.global
	},
	alsoletter = {.},
	sensitive = false,
	morecomment = [f][\em\color{comment}][0]{*},
	morecomment = [l]{;} 
}

\lstdefinestyle{Verilog}{
	language			= Verilog,
    basicstyle          = {\footnotesize\ttfamily},
    belowskip           = 0pt,
    columns             = fixed,
    captionpos          = b,
	numbers				= left,
	numberstyle			= \color{number},
	numbersep			= 5pt,
	frame				= tb,
	xleftmargin			= 1.5em,
	framexleftmargin	= 1.5em,
	tabsize				= 2,
	showtabs			= false,
	showspaces			= false,
	showstringspaces	= false,
    keepspaces          = true,
    breaklines          = true,
	commentstyle		= \color{comment},
	keywordstyle		= \color{keyword}\bfseries,
	stringstyle			= \color{string},
	morekeywords		= {
		analog, branch
	},
	moredelim			= [is][\color{constant}\bfseries]{(C)}{(C)},
	moredelim			= [is][\color{macro}\bfseries]{(M)}{(M)},
	moredelim			= [is][\color{red}\bfseries]{(P)}{(P)},
	moredelim			= [is][\color{blue}\bfseries]{(F)}{(F)},
	keywordstyle		= [2]\color{discipline}\bfseries,
	keywords			= [2]{
		electrical, rotational, rotational\_omega, ground
	},
	keywordstyle		= [3]\color{keyword}\bfseries,
	keywords			= [3]{
		ddt, idt, log, pow, cos, sin, tanh, tanhsw, min, max, laplace\_nd, from
	},
	keywordstyle		= [4]\color{red}\bfseries,
	keywords			= [4]{V, Theta, Omega},
	keywordstyle		= [6]\color{blue}\bfseries,
	keywords			= [6]{I, Tau}
}

\usepackage{glossaries}

\newacronym{auc-roc}{AUC-ROC}{Area Under the Receiver Operating Characteristic Curve}
\newacronym{aadl}{AADL}{Architecture Analysis \& Design Language}
\newacronym{ac}{AC}{Automation Controller}
\newacronym{adc}{ADC}{Analog-to-Digital Converter}
\newacronym{ag}{A/G}{Assume-Guarantee}
\newacronym{agv}{AGV}{Automated Guided Vehicle}
\newacronym{ams}{AMS}{Analog-Mixed Signals}
\newacronym{api}{API}{Application Programming Interface}
\newacronym{aml}{AutomationML}{Automation Markup Language}
\newacronym{amqp}{AMQP}{Advanced Message Queuing Protocol}
\newacronym{ab}{AB}{AdaBoost}
\newacronym{ae}{AE}{autoencoder}
\newacronym{arlstm}{AR-LSTM}{Autoregressive Long Short-Term Memory}
 
\newacronym{bom}{BOM}{Bill of Materials}
\newacronym{bdd}{BDD}{Block Definition Diagram}
\newacronym{elbo}{ELBO}{Evidence Lower BOund}

\newacronym{cnc}{CNC}{Computerized Numerical Control}
\newacronym{cnn}{CNN}{Convolutional Neural Network}
\newacronym{csp}{CSP}{Constraint Satisfaction Programming}
\newacronym{cad}{CAD}{Computer-aided design}
\newacronym{casse}{CASSE}{Communication Aware Specification and Synthesis
	Environment}
\newacronym{cps}{CPS}{Cyber-Physical System}
\newacronym{cpps}{CPPS}{Cyber-Physical Production System}

\newacronym{caex}{CAEX}{Computer Aided Engineering Exchange}
\newacronym{collada}{COLLADA}{COLLAborative Design Activity}
\newacronym{cnnvae}{CNN-VAE}{Convolutional variational autoencoders}

\newacronym{dse}{DSE}{Design Space Exploration}
\newacronym{dac}{DAC}{Digital-to-Analog Converter}
\newacronym{dsl}{DSL}{Domain Specification Language}
\newacronym{din}{DIN}{Deutsches Institut f\"ur Normung}
\newacronym{dih}{DIH}{Data Integration HUB}
\newacronym{b2mml}{B2MML}{Business To Manufacturing Markup Language}
\newacronym{dfjss}{DFJSS}{Dynamic Flexible Job Shop Scheduling}
\newacronym{dof}{DoF}{Degree of Freedom}
\newacronym{dl}{DL}{Deep Learning}

\newacronym{eda}{EDA}{Electronic Design Automation}
\newacronym[longplural=Electromagnetic Interferences]{emi}{EMI}{Electromagnetic Interference}
\newacronym{esl}{ESL}{Electronic System Level}
\newacronym{eln}{ELN}{Electrical Linear Network}
\newacronym{erp}{ERP}{Enterprise Resource Planning}

\newacronym{fmi}{FMI}{Functional Mock-up Interface}
\newacronym{fms}{FMS}{Flexible Manufacturing System}
\newacronym{fmu}{FMU}{Functional Mock-up Unit}
\newacronym{fmea}{FMEA}{Failure Mode and Effect Analysis}
\newacronym{fsm}{FSM}{Finite State Machine}
\newacronym{fdm}{FDM}{Fused Deposition Modeling}
\newacronym{fjss}{FJSS}{Flexible Job Shop Scheduling}

\newacronym{gr1}{GR(1)}{General Reactivity}
\newacronym{gbrf}{GBRF}{Gradient Boosted Regression Forest}

\newacronym{hdl}{HDL}{Hardware Description Language}
\newacronym{hif}{HIF}{Heterogeneous Intermediate Format}
\newacronym{hrm}{HRM}{Hardware Resource Modeling}
\newacronym{hw}{HW}{Hardware}
\newacronym{hmi}{HMI}{Human-Machine Interaction}

\newacronym{ic}{IC}{Integrated Circuit}
\newacronym{ice}{ICE}{Industrial Computer Engineering}
\newacronym{ios}{IoS}{Internet of Services}
\newacronym{iot}{IoT}{Internet of Things}
\newacronym{iiot}{IIoT}{Industrial Internet of Things}
\newacronym{isa}{ISA}{International Society of Automation}
\newacronym{iss}{ISS}{Instruction Set Simulator}
\newacronym[longplural={Intellectual Properties}]{ip}{IP}{Intellectual Property}
\newacronym{ibd}{IBD}{Internal Block Diagram}
\newacronym{imu}{IMU}{Inertial Measurement Unit}

\newacronym{jss}{JSS}{Job Shop Scheduling}

\newacronym{kpn}{KPN}{Khan Process Networks}
\newacronym{knn}{kNN}{k-Nearest Neighbors}
\newacronym{ltl}{LTL}{Linear Temporal Logic}
\newacronym{lstm}{LSTM}{Long Short-Term Memory}
\newacronym{mbd}{MBD}{Model-based Design}
\newacronym{mbse}{MBSE}{Model-based System Engineering}
\newacronym{mems}{MEMS}{Micro Electro Mechanical Systems}
\newacronym{milp}{MILP}{Mixed Integer Linear Programming}
\newacronym[longplural={Models of Computation}]{moc}{MoC}{Model of Computation}
\newacronym{mu}{MU}{Mobile Unit}
\newacronym{mes}{MES}{Manufacturing Execution System}
\newacronym{mom}{MOM}{Manufacturing Operations Management}
\newacronym{m2m}{M2M}{Machine to Machine}
\newacronym{ml}{ML}{Machine Learning}
\newacronym{mrmr}{mRMR}{Minimum Redundancy Maximum Relevance}
\newacronym{mlp}{MLP}{Multilayer Perceptron}
\newacronym{mlpb}{MLP-B}{Bayesian Multilayer Perceptron}
\newacronym{msr}{MSR}{Maximum Softmax Response}
\newacronym{mtsad}{MTSAD}{Multivariate Time Series Anomaly Detection}
\newacronym{mts}{MTS}{Multivariate Time Series}

\newacronym{nes}{NES}{Networked Embedded System}
\newacronym{noc}{NoC}{Network on Chip}
\newacronym[longplural={Non-Functional-Properties}]{nfp}{NFP}{Non-Functional-Property}
\newacronym{ntp}{NTP}{Network Time Protocol}

\newacronym{opcua}{OPC~UA}{OPC Unified Architecture}
\newacronym{ostc}{O$S^3$TC}{Open-Source Smart-System Test Case}
\newacronym{ovp}{OVP}{Open Virtual Platform}

\newacronym{pbd}{PBD}{Platform-Based Design}
\newacronym{plc}{PLC}{Programmable Logic Controller}
\newacronym{ppe}{PPE}{Personal Protective Equipment}
\newacronym{ptp}{PTP}{Precision Time Protocol}
\newacronym{pc}{PC}{Personal Computer}

\newacronym{qc}{QC}{Quality Checking}

\newacronym{roc}{ROC}{Receiver Operating Characteristic}
\newacronym{rtl}{RTL}{Register-Transfer Level}
\newacronym{rmse}{RMSE}{Root Mean Square Error}
\newacronym{road}{RoAD}{Robotic Arm Dataset}
\newacronym{rpc}{RPC}{Remote Procedure Call}
\newacronym{rtn}{RTN}{Resource Task Network}
\newacronym{rf}{RF}{RandomForest}
\newacronym{rtt}{RTT}{Round Trip Time}
\newacronym{sdk}{SDK}{Software Development Kit}
\newacronym{sld}{SLD}{System-Level Design}
\newacronym{soc}{SoC}{System on a Chip}
\newacronym{srm}{SRM}{Software Resource Modeling}
\newacronym{sw}{SW}{Software}
\newacronym{sme}{SME}{Small and Medium Enterprise}
\newacronym{sysml}{SysML}{System Modeling Language}
\newacronym{scnsl}{SCNSL}{SystemC Network Simulation Library}
\newacronym{soa}{SOA}{Service Oriented Architecture}
\newacronym{som}{SOM}{Service Oriented Manufacturing}
\newacronym{scada}{SCADA}{Supervisory Control and Data Acquisition}
\newacronym{stomp}{STOMP}{Streaming Text
	Oriented Messaging Protocol}
\newacronym{stn}{STN}{State Task Network}
\newacronym{svm}{SVM}{Support Vector Machine}
\newacronym{ssl}{SSL}{Self-Supervised Learning}

\newacronym{tlm}{TLM}{Transaction-Level Modeling}

\newacronym{uml}{UML}{Unified Modeling Language}

\newacronym{vlsi}{VLSI}{Very Large Scale Integration}
\newacronym{vp}{VP}{Virtual Platform}

\newacronym{wsn}{WSN}{Wireless Sensor Networks}

\newacronym{xmi}{XMI}{XML Metadata Interchange}

\newcommand*{\toolname}{VARADE}

\copyrightyear{2024}
\acmYear{2024}
\setcopyright{acmcopyright}
\acmConference[DAC '24]{61st ACM/IEEE Design Automation Conference}{June 23--27, 2024}{San Francisco, CA, USA}
\acmBooktitle{61st ACM/IEEE Design Automation Conference (DAC '24), June 23--27, 2024, San Francisco, CA, USA}
\acmDOI{10.1145/3649329.3655691}
\acmISBN{}

\begin{document}

\title{\toolname: a Variational-based AutoRegressive model for Anomaly Detection on the Edge}

\author{Alessio Mascolini$^1$, Sebastiano Gaiardelli$^2$, Francesco Ponzio$^1$, Nicola Dall'Ora$^2$, Enrico Macii$^1$,\ \ \ \ Sara Vinco$^1$, Santa {Di~Cataldo}$^1$, Franco Fummi$^2$}
\affiliation{%
 \institution{%
    $^1$Politecnico di Torino, Turin, Italy, \{name.surname\}@polito.it\\
    $^2$University of Verona, Verona, Italy, \{name.surname\}@univr.it}%
    \country{}
}

\renewcommand{\shortauthors}{Mascolini Alessio, Sebastiano Gaiardelli, Francesco Ponzio, Nicola Dall'Ora, et al.}

\begin{abstract}
Detecting complex anomalies on massive amounts of data is a crucial task in Industry 4.0, best addressed by deep learning. However, available solutions are computationally demanding, requiring cloud architectures prone to latency and bandwidth issues. This work presents \toolname{}, a novel solution implementing a light autoregressive framework based on variational inference, which is best suited for real-time execution on the edge. The proposed approach was validated on a robotic arm, part of a pilot production line, and compared with several state-of-the-art algorithms, obtaining the best trade-off between anomaly detection accuracy, power consumption and inference frequency on two different edge platforms.
\end{abstract}


\maketitle

\begin{acks}
The work has been partially supported by PRIN 2022T7YSHJ SMART-IC - Next Generation EU project. This manuscript reflects only the Authors’ views and opinions, neither the European Union nor the European Commission can be considered responsible for them.
\end{acks}

\section{Introduction}
\label{sec:introduction}
In any production context, the downtime of a machine due to the sudden breakdown of mechanical, hydraulic, or electrical components leads to severe losses in terms of time and money. For this reason, efforts are spent for the 
early detection of any irregular behavior of the production line to avoid sudden stops, enable specific preventive maintenance actions, and reduce the environmental impact. This evolution is enabled by transforming traditional production machinery into \glspl{cps}, where sensor devices, communication technologies, and data analytics cooperate to manage production failures in advance~\cite{SUVARNA20211212}. 

In an industrial \gls{cps} scenario, the most crucial resource is the availability of data reflecting the different aspects of production. Such data consist of multiple interdependent variables rapidly evolving over time, thus falling under the typical definition of \emph{\gls{mts}}~\cite{Li2023}. After collection, the time series, originated by heterogeneous sensors and data sources, are integrated through \gls{iiot} technologies and made available for anomaly detection, visualization, and analysis~\cite{Yang2023}.

Although extensive research has been carried out on \gls{mtsad}, current
solutions typically lack the flexibility and scalability that is required for an
effective real-time deployment~\cite{nain2022towards, yu2022edge}. 
In most proposed solutions, the raw data are in fact streamed through the
\gls{iiot} network to a cloud platform~\cite{s22072445}, where an expert
data-driven system is in charge of the anomaly detection stage. This typically
results in high latency owed to communication overhead~\cite{Bacchiani2022}.
Unlike \gls{iot} networks, \gls{iiot} networks are characterized by sensors transmitting a massive amount of data that must be processed in real-time~\cite{HU2019569, qiu_edge_2020}.
The order of magnitude of the transmitted data can be GB/s for large production 
plants, making cloud processing impracticable due to the bandwidth requirements and 
impairments~\cite{qiu_edge_2020}.
All such
considerations highlight that anomalies should be detected \emph{as soon as
possible} and \emph{as close as possible} to the monitored \gls{cps}, preferably
with real-time or near real-time response, rather than on the
cloud. This makes \emph{edge computing} strategic to maintain the overall system
functionally safe~\cite{Li2023, DeVita2023}. 

On top of these considerations, in our study, we propose \toolname, a novel real-time and \emph{edge-friendly} anomaly detection solution that provides a new efficient training paradigm for \emph{light} \gls{mtsad}. 
The autoregressive framework of \toolname\ allows handling streaming data with minimal latency. Furthermore, its variational formulation obtains the best compromise between model compactness and anomaly detection accuracy, making it best suited for real-time execution on the edge.

To prove the effectiveness of \toolname\ in a real industrial \gls{cps} scenario, we employ a collaborative robot working in a fully-fledged manufacturing line and providing a continuous stream of heterogeneous sensor data. On this testbed, we compare \toolname\ with a comprehensive set of state-of-the-art \emph{light} (i.e., edge-suitable) \gls{mtsad} solutions on two different edge platforms. 
Our experiments demonstrate that \toolname{} performs well even with limited computational resources, with an optimal balance between required power, anomaly detection accuracy, and inference frequency of the model, which can be varied according to the industrial machinery being monitored.


This paper is organized as follows. Section~\ref{sec:background} presents the necessary background and related works. Section~\ref{sec:methods} details the proposed \toolname\ anomaly detection method, as well as the other benchmarked solutions. Section~\ref{sec:discussion} describes the collaborative robot case-study and discusses the experimental results. Finally, \sectionautorefname~\ref{sec:conclusions} draws our concluding remarks.

\section{Background and related works}
\label{sec:background}
\gls{mtsad} scenarios are best addressed with \gls{dl} methodologies, that recently proved to be more effective in tackling complex anomalies in \gls{mts} data 
\cite{schmidl2022anomaly, li2022dl_ad_review} than traditional anomaly detection methods (e.g., based on clustering or statistical indexes~\cite{zhou2022contrastive}). 
Nonetheless, most \gls{dl}-based solutions present major drawbacks in terms of required data transmission and/or high computational cost~\cite{yu2022edge,liu2023efficient}.
On the other hand, \emph{light} \gls{mtsad} models compatible with edge computing are typically based on tiny and scaled \glspl{cnn}, that need to be trained with huge sets of annotated anomalies~\cite{yu2022edge, nain2022towards, sun2023tinyad, liu2023efficient}. Collecting and annotating such training sets is however unfeasible in most industrial applications~\cite{nain2022towards,li2022dl_ad_review}.

To circumvent this problem, the most promising approach is to learn the characteristics of a ``normal'' behavior from a large amount of non-anomalous data so as to be able to identify any events that significantly deviate from the normality, with three different strategies: i)~forecasting-based, ii)~reconstruction-based, and iii)~outlier detection methods. 

\emph{Forecasting-based} methods learn to predict a number of time steps leveraging a current context window. Then, they compare the predicted values with the observed ones to identify anomalies~\cite{schmidl2022anomaly}. A large number of studies in this group employ autoregressive \gls{lstm} networks, a type of recurrent network able to learn long-term time dependencies in multivariate data~\cite{cook2019anomaly, kumar2023analysis, lu2023review}. A recent work leverages instead a forest of gradient boosted regression trees to detect anomalies in a Digital Twin-driven industrial context, by examining the residuals from the forecasts of an ensemble of weak predictors~\cite{GBRF}. 

\emph{Reconstruction-based} methods encode the characteristics of a normal time series into a latent representation and learn to reconstruct new data starting from it. The reconstruction error is then exploited to discriminate the anomalous values from the normal ones. The most popular methods in this group are built on top of \glspl{ae}, encoder-decoder neural networks where the encoder learns a compressed version of the input data, and the decoder learns to recreate the input starting from the encoded representation. Among the others, \cite{kim2018squeezed} employed convolutional \glspl{ae} for anomaly detection in an IoT-inspired environment, and proved that reducing the size, complexity, and training cost of the \gls{ae} did not lower its ability to identify anomalies.

\begin{figure}[!hbt]
    \centering
    \includegraphics[width=.85\columnwidth]{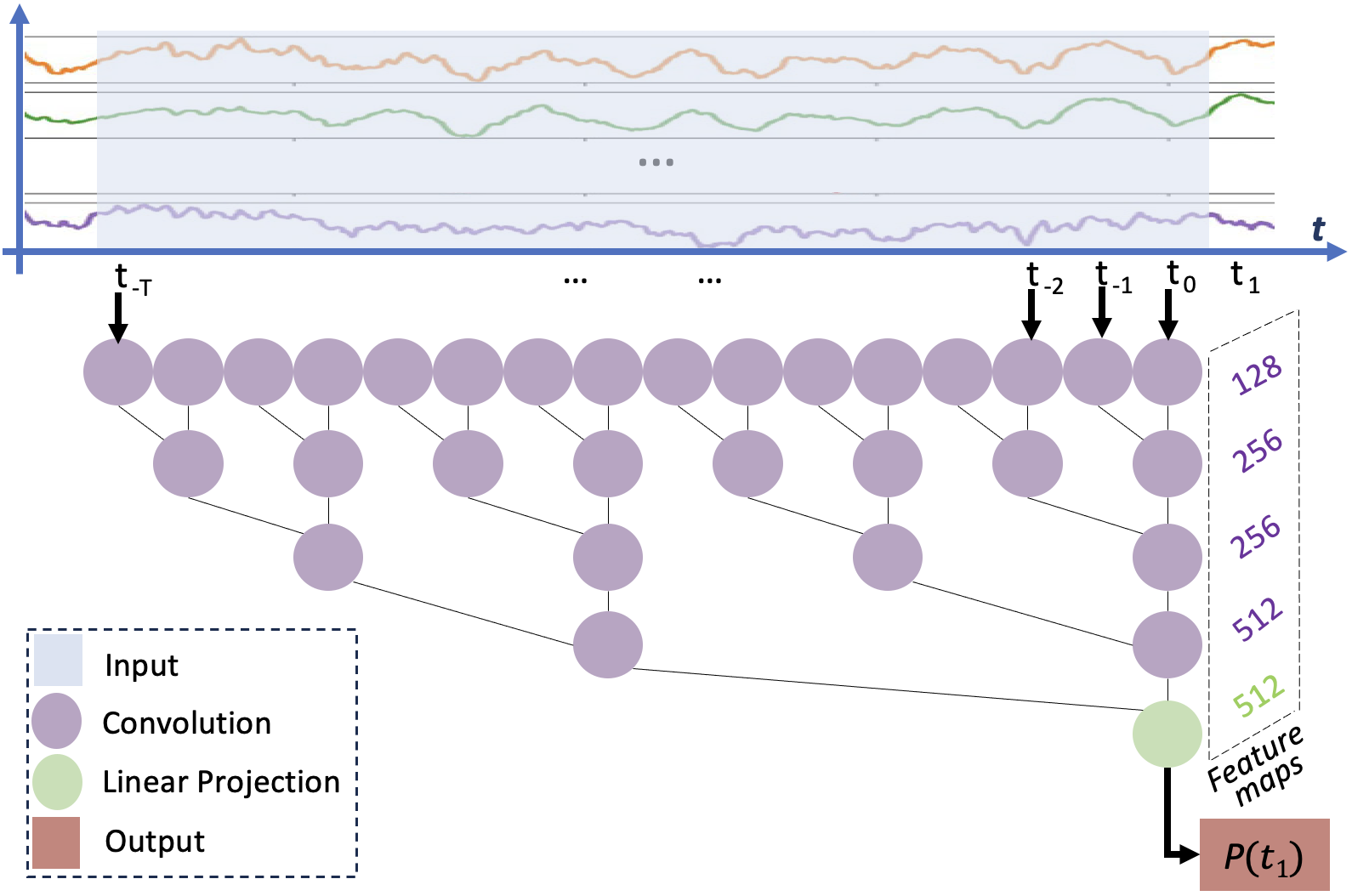}
    \caption{Architecture of \toolname. Current ($t_0$) and past time steps ($t_{-1} \dots t_{-T}$) are processed by a cascade of convolutional layers and a final linear projection. The output is the estimated probability distribution of the next time step, $P(t_{1})$.}
    \vspace{-1.5em}
    \label{fig:architecture}
\end{figure}

\emph{Outlier detectors} identify anomalies based on their dissimilarity from regular data points in the feature space. Popular edge-friendly examples in this group are based on \gls{knn}, identifying anomalous values based on the distance from their neighbours, and Isolation Forest, that uses the number of binary splits necessary for an ensemble of decision trees to isolate the point from the rest of the data
~\cite{liu2012isolation}.

\section{Methods}
\label{sec:methods}
\subsection{Proposed solution}

\toolname\ works to strike a balance between traditional techniques, that offer quick inference but with limited accuracy, and \gls{dl} models, that learn complex patterns but require substantial computational resources, not available at the edge. 
Our design choice is to employ a \emph{forecasting-based autoregressive framework}: by predicting samples one at a time based on previous ones, this framework is naturally suited to handle streaming data with minimal latency.

\figureautorefname~\ref{fig:architecture} illustrates in principle the proposed architecture. The model takes as input the samples at the current ($t_0$) and past time steps ($t_{-1}$,...$t_{-T}$, with $T=16$ for visualization purposes), and passes them through a set of convolutional layers with ReLU activations and a linear projection, to finally predict a single future time step, $t_{1}$. The reason behind this architectural choice lies in the consideration that 
model inference speed of \glspl{cnn} is commonly limited by memory bandwidth, especially with SIMD implementations for CPUs and CUDA kernels for GPUs~\cite{ghimire2022survey}. By using convolutions with kernel size and stride of 2, we obtain that the time-dimension is halved at every new layer, leading to very limited memory usage and bandwidth requirements compared to the number of parameters, and hence to faster inference. On the other hand, the number of feature maps is doubled every two layers (see \figureautorefname~\ref{fig:architecture}), helping the network to learn more complex and abstract features.

Conventional forecasting-based anomaly detectors work by considering an \emph{anomaly score}, measured as the euclidean norm between the forecasted value and the measured one. 
In our experiments, we have observed that a \gls{dl}-based autoregressive model compact enough for real-time execution on edge fails to deliver satisfactory forecasting performance, dramatically affecting the quality of the anomaly scores. This lead us to a probabilistic approach, where the model outputs a probability distribution of the possible values for the next data point in the sequence ($P(t_{1})$ in \figureautorefname~\ref{fig:architecture}).


Predicting a probability distribution using a neural network is a complex problem, which can be greatly simplified by constraining the distribution to be Gaussian. This approach, known as \emph{variational inference}~\cite{blei2017variational}, leads to a simpler optimization problem where the objective is to find the mean and variance which minimize the loss function (details will follow). The additional advantage of the Gaussian constraint is that the variance can be interpreted as the uncertainty of the prediction: since we expect the model to be more confident in its prediction when the system is operating normally, and less confident when an anomaly is occurring, the variance can be directly used as an \emph{anomaly score}.

Summarizing, the proposed architecture is composed by $N$ convolutional layers, with time-dimension halved at every layer. Hence, $N$ strictly depends on the input window size $T$. In our work, we set $T=512$, resulting into a total of $8$ layers. Conversely, the number of feature maps is doubled every two layers starting from $128$, which leads to $1,024$ in the final layer. The output mean and variance values of the estimated probability distribution $P(t_{1})$ are obtained at last by linear projection.

\subsection{Derivation of the loss function}
As loss function we employ the inverse of the 
\gls{elbo}, that provides a lower bound on the log evidence, $log\,p(x)$, where $x$ represents the observed data. Thus, maximizing the \gls{elbo} leads to a better approximation of the true posterior.

The \gls{elbo} can be decomposed into two terms:
\begin{itemize}
    \item The expectation of the log-likelihood under the approximate posterior, which pushes the approximate distribution to put more probability mass on configurations of the latent variables that explain the observed data well.

    \item The negative divergence between the approximate and prior distribution, which encourages the approximate distribution to be close to the prior.
\end{itemize}

By presuming a Gaussian distribution, our model predicts both the mean and the logarithm of the distribution's variance. We opt for the logarithm over the simple variance, as the latter can only be positive. Hence, the reconstruction loss is essentially computing the negative log-likelihood of the observed data under a Gaussian distribution assumption.

Let us assume that our data \(y\) is normally distributed with a mean of \(\mu\) and a variance of \(\sigma^2\). The probability density function (PDF) of a normal distribution is given by:
\begin{equation}
\small
p(y|\mu, \sigma^2) = \frac{1}{\sqrt{2\pi\sigma^2}} \exp \left(-\frac{(y-\mu)^2}{2\sigma^2}\right) 
\end{equation}

Taking the negative logarithm of the PDF to get the negative log-likelihood (NLL), we have:
\begin{equation}
\small
NLL(y|\mu, \sigma^2) = - \log \left(\frac{1}{\sqrt{2\pi\sigma^2}} \exp \left(-\frac{(y-\mu)^2}{2\sigma^2}\right)\right) 
\end{equation}

After simplifying the above equation, we get:
\begin{equation}
\small
NLL(y|\mu, \sigma^2) = \frac{1}{2} \log(2\pi\sigma^2) + \frac{(y-\mu)^2}{2\sigma^2}
\end{equation}

Given that \(\log(2\pi)\) is just a constant, we can ignore it during optimization (as it depends on the derivative of the loss, and the derivative of a constant is zero). This simplifies to:
\begin{equation}
\small
NLL(y|\mu, \sigma^2) = \frac{1}{2} \log(\sigma^2) + \frac{(y-\mu)^2}{2\sigma^2} 
\end{equation}

So, in our case, the reconstruction loss we use is:
\begin{equation}
\small
L_{\text{recon}} = \frac{1}{2} \left( \log(\sigma_{\text{pred},i}^{2}) + \frac{(y_{i} - \mu_{\text{pred},i})^{2}}{\sigma_{\text{pred},i}^{2}} \right) 
\end{equation}
where \( \sigma_{\text{pred}}^{2} \) represents the predicted variance, \( \mu_{\text{pred}} \) the predicted mean and \(i\) the current step. This formula encourages our model to predict a distribution close to the actual data.

The next part of the loss calculation introduces the Kullback–Leibler (KL) divergence, which quantifies the difference between our predicted distribution and our prior, a standard Gaussian distribution. This is computed as:
\begin{equation}
\small
D_{KL} = -\frac{1}{2} \left( 1 + \log(\sigma_{\text{pred}}^{2}) - \mu_{\text{pred}}^{2} - \sigma_{\text{pred}}^{2} \right) 
\end{equation}
The $D_{KL}$ term encourages the model to predict the data mean and variance when it is uncertain. This helps regularize our model and is critical to employ our anomaly detection method.

The final loss function \( L \) is a weighted sum of the reconstruction loss \( L_{\text{recon}} \) and the KL divergence \( D_{KL} \):
\begin{equation}
\small
L = L_{\text{recon}} + \lambda D_{KL} 
\end{equation}

Thanks to the KL divergence term, our model learns to predict a higher variance when it is uncertain about the next value, and a low variance when it is confident. During inference, the mean prediction is removed, and the variance is directly used as an \emph{anomaly score}: the higher the score, the larger the detected anomaly.

\subsection{Baseline solutions}
\label{ssec:baseline_methods}

As a baseline for our proposed method, in this study, we implement and analyze a representative sample of \emph{light} anomaly detectors that have been successfully deployed in edge computing scenarios, by considering the approaches in \sectionautorefname~\ref{sec:background}. 
\begin{itemize}
  \item \emph{\gls{arlstm}}. A recurrent architecture featuring 5 \gls{lstm} recurrent layers with 256 feature maps each, followed by 2 fully connected layers. The anomaly score is then calculated as the euclidean norm of the difference between predicted and real value, as in many previous works~\cite{lstmRoy, cook2019anomaly, kumar2023analysis, lu2023review}. To find the best configuration for our specific task, we follow the memory-efficient paradigm introduced by~\cite{dlstm}, and pick a number of layers equal to 5 based on past experiments at a similar window size~\cite{lstmTraffic}. 
  \item \emph{\gls{gbrf}}. The technique presented in \cite{GBRF}, with minor modifications to boost the anomaly detection capabilities: the number of decision trees is increased from 5 to 30 and the dimensionality reduction step is removed. The anomaly score is computed in the same way as for \gls{arlstm}.
        \item \emph{Autoencoder (\gls{ae})}. A convolutional autoencoder featuring 6 ResNet blocks \cite{resnet}. 
The anomaly score is the euclidean norm of the difference of reconstructed and real value.
        \item \emph{\gls{knn}}. Past works show \gls{knn} as the best performing nearest neighbour based algorithm for anomaly detection, with anomaly score computed either as the average or the maximum distance from the neighbors \cite{knnAD}. We employ maximum distance with k=5, as it has the best compromise between accuracy and execution time. 
        \item \emph{Isolation Forest}. An ensemble of 100 individual decision trees, that isolate each data point into a leaf. The anomaly score of a data point is based on the average path length \cite{liu2012isolation}. As recommended by \cite{liu2012isolation}, we use a contamination value of 0.1, which defines the proportion of outliers in the dataset.
    \end{itemize}

\subsection{Implementation details}
All the models were implemented in TensorFlow 2.11.0 and Sklearn 1.1.2. For a fair comparison, all the anomaly detection frameworks were trained in the same experimental conditions and implementing 
hyperparameters tuning strategies \footnote{Source code and dataset available at https://gitlab.com/AlessioMascolini/varade}. More specifically: the neural network-based frameworks were optimized using Adam with a fixed $10^{-5}$ learning rate. GBRF and Isolation Forest were trained using the mean squared error criterion and recursive binary splitting, by strictly following the respective reference papers.

\section{Industrial case study}
\label{sec:discussion}
\subsection{Kuka anthropomorphic manipulator}

To create a realistic scenario for the anomaly detection methods, we focused our case study on a KUKA LBR iiwa collaborative industrial robot, part of a fully-fledged production line\footnote{Industrial Computer Engineering Laboratory - https://www.icelab.di.univr.it/}. 
The robot performs pick and place operations and it is controlled by a Simatic S7-1200 \gls{plc}, 
directly connected to the robot through a hard-wired field bus. 
The \gls{plc} runs an \gls{opcua} server, that exposes the KUKA state and functionality as services: the activation of such services in a given order constitutes a production process. 

The KUKA robot allows collecting the robot's parameters 
through its programming interface. However, this limits the 
frequency with which such parameters can be collected to 5 Hz. At higher frequencies, queries interfere with the controlling 
process, causing stuttering in the robot trajectories.  
For this reason, we instrumented the KUKA robot with seven \gls{imu} sensors (DFRobot SEN0386), one on each robot joint, to  measure the joint's angle, acceleration, 
and angular velocity. 
These sensors send data at 200 Hz on a serial wire after applying a Kalman filter to reduce noise. 
In addition to physical data, we collected also extra-functional data from a single-phase energy meter (Eastron SDM230) monitoring the energy consumption of both the robot 
and the industrial PC. 
This energy meter is connected through a hard-wired Modbus with an industrial ESP-32 
(Olimex ESP32-EVB), collecting and sending data to a MQTT broker via Ethernet.

\begin{figure}[!t]
    \centering
    \includegraphics[width=0.75\columnwidth]{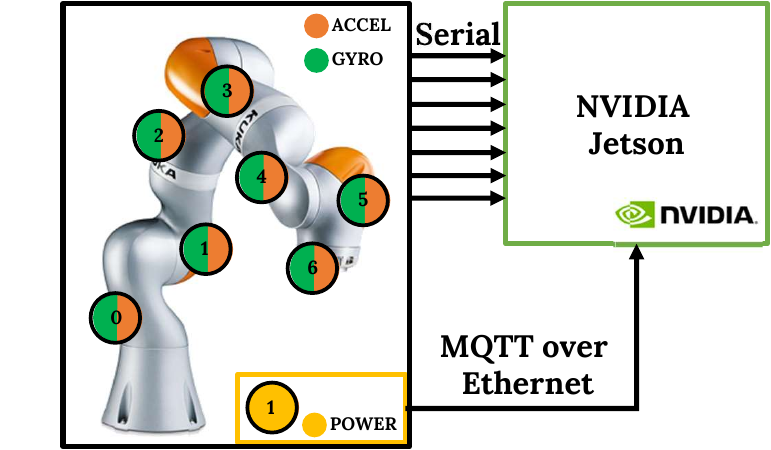}
    \caption{Case study setup. The KUKA manipulator is instrumented with different sensors: 7 accelerometers with six axes (one for each joint) and one single-phase power meter. The sensors are connected directly with an embedded board for detecting different classes of anomalies.}
    \vspace{-1.2em}
    \label{fig:case_study}
\end{figure}

\cref{fig:case_study} depicts the experimental setup, consisting of: the KUKA robot,
seven \gls{imu} sensors, an energy meter, and an embedded board connected to the sensors and executing the 
anomaly detection model (further described 
in \cref{ssec:experiamental_results}). 

\subsection{Data stream characterization}

The data stream collected from  the robotic manipulator consist of 86 channels in total
(reported in \cref{tab:datastream}),
including signals to monitor the action currently performed by the robot (\ie{}, \emph{action ID}), its kinematic behavior (\emph{Joint Channels}) and its 
extra-functional parameters (\emph{Power Channels})~\cite{mascolini2023robotic}.

The Joint Channels consist of data related to the seven joints collected from the \glspl{imu}
sensors, each having the same eleven components that monitor various aspects of motion and temperature.
Originally, the \gls{imu} collect angles in the $[-180,+180]~^{\circ}C$ range, causing high value changes when rotating near the two extremes. Since this may be a source of confusion for pattern recognition techniques, we had the orientations converted to quaternions, a 4-dimensional
coordinate system commonly used in robotics.

The Power channels consist of eight quantities monitored by the energy meter. 
These channels allow detecting anomalies that could be transparent with respect to the robot trajectories, such as high power draw from a motor. 

\begin{table}[!b]
    \centering
    \vspace{-0.5em}
    \caption{Channels description: for each sensor considered, the related variables are listed. The  <X> in the variable name is a label representing the [0,6] index of the joint in which the corresponding \gls{imu} sensor is placed on the robot.}
    \vspace{-0.5em}
    \label{tab:datastream}
    \resizebox{.85\linewidth}{!}{
        \footnotesize
    \begin{tblr}{
  width = \linewidth,
  colspec = {Q[52]Q[292]Q[162]Q[390]},
  column{3} = {c},
  cell{1}{2,4} = {c},
  cell{3}{1} = {r=11}{},
  cell{14}{1} = {r=7}{},
  vline{3-4} = {1-2,4-9,11-20}{},
  vline{3} = {2}{-}{},
  vline{2-4} = {1-9,10-20}{},
  hline{2-3,14,21} = {-}{},
  hline{2} = {2}{1-4}{},
}
 & \textbf{Channel name} & \textbf{Unit} & \textbf{Description}\\
 & action ID & - & Robot action ID\\
 \begin{sideways}\emph{Joint Channels}\end{sideways} & sensor\_id\_X\_AccX & m/s$^2$ & X-axis acceleration\\
 & sensor\_id\_X\_AccY & m/s$^2$ & Y-axis acceleration\\
 & sensor\_id\_X\_AccZ & m/s$^2$ & Z-axis acceleration\\
 & sensor\_id\_X\_GyroX & deg/s & X-axis angular velocity\\
 & sensor\_id\_X\_GyroY & deg/s & Y-axis angular velocity\\
 & sensor\_id\_X\_GyroZ & deg/s & Z-axis angular velocity\\
 & sensor\_id\_X\_q1 & - & Quaternion orient. comp. 1\\
 & sensor\_id\_X\_q2 & - & Quaternion orient. comp. 2\\
 & sensor\_id\_X\_q3 & - & Quaternion orient. comp. 3\\
 & sensor\_id\_X\_q4 & - & Quaternion orient. comp. 4\\
 & sensor\_id\_X\_temp & $^\circ$C & Temperature\\
\begin{sideways}\emph{Power Channels}\end{sideways} & current & A & Current\\
 & frequency & Hz & Frequency\\
 & phase\_angle & degree & Phase angle\\
 & power & W & Power\\
 & power\_factor & - & Power factor\\
 & reactive\_power & VAr & Reactive power\\
 & voltage & V & Voltage
\end{tblr} 
    }
    \vspace{-1.0em}
    
\end{table}

\subsection{Experimental setup}

To train \toolname\ and the baseline anomaly detection models, we 
created a dataset by recording the robot performing 30 unique actions (\ie{}, its machine services) executed in a cycle for a total duration of 390 minutes. 
The resulting dataset contains all the possible actions supported by the robot, distributed uniformly within its duration. 
This allows the offline training of the anomaly detection models on the \emph{``normal behavior''} 
of the robot in all the possible production processes supported by the manufacturing system. 
Given the diverse nature of the anomaly detection models, the collected data are normalized in the range [-1, 1] based on the minimum and maximum values of each sensor's data, ensuring that all the features have equal importance avoiding unfair comparison.  

To test the trained models in real-time conditions, we designed a "collision experiment" of 82 minutes in total. During this experiment, the robot performed all the 30 possible actions. During the robot operations, 125 collision anomalies were randomly generated by a human operator, by manually interfering 
with the robot during its movement in a very limited timeframe.
This simulates sudden collisions between a human worker (or an object) and the robot, which is a realistic hazardous situation in a production line. 

To test the suitability to an edge scenario, we selected two edge devices, connected to the robotic system depicted in \cref{fig:case_study}: 
a Nvidia Jetson Xavier NX
(with 6 cores and 16 GB of RAM) and a Jetson AGX Orin (with 12 cores and 32 GB of RAM).
Each anomaly detection model has been tested by a software script that continuously reads data from the sensors, prepares the data by applying a preprocessing function, and calls the inference function.

During each test, the anomaly detection accuracy was evaluated in terms of \gls{auc-roc} value. The ratio is to interpret an anomaly detector as a binary classifier, where points are classified as anomalous if the anomaly score exceeds a certain threshold. The \gls{roc} curve plots the true positive
rate against the false positive rate at varying values of this threshold,
and the area under this curve provides a single threshold-less
$[0,1]$ measure of the algorithm’s ability to identify the anomalous
data points. 

Besides the \gls{auc-roc} score, we measured the inference frequency, and we collected all the most relevant board's metrics (\emph{e.g.}, power consumption, RAM usage,
GPU RAM usage) by exploiting the \textit{jetson-stats} library.
These metrics were collected not only during the execution of the anomaly detection tasks, but also with the boards in \emph{Idle state} for 6 minutes: the mean value was computed as baseline to evaluate the load introduced by the anomaly detection models \wrt\ the standard state.

\subsection{Experimental results}
\label{ssec:experiamental_results}


\cref{tab:jetson-results} reports the results obtained by \toolname{} and by the baseline detectors described in \cref{ssec:baseline_methods} on the two selected edge devices.

\begin{table*}[hbt]
    \centering
    \caption{Comparison between the Anomaly Detection models executed in real-time on the two edge processing units. 
    }
    \label{tab:jetson-results}
    \vspace{-.7em}
    \resizebox{.85\linewidth}{!}{
        \footnotesize
        \begin{tabular}{c|c|| d{2.3} | d{2.3} | d{4.3} | d{3.3} | d{2.3} | d{1.3} | d{2.3} } 
\begin{tabular}[c]{@{}c@{}}Board \\ Model\end{tabular} &
  \begin{tabular}[c]{@{}c@{}}Anomaly Detection \\ Model\end{tabular} &
  \multicolumn{1}{c|}{\begin{tabular}[c]{@{}c@{}}CPU Usage\\ (\%)\end{tabular}} &
  \multicolumn{1}{c|}{\begin{tabular}[c]{@{}c@{}}GPU Usage\\ (\%)\end{tabular}} &
  \multicolumn{1}{c|}{\begin{tabular}[c]{@{}c@{}}RAM Usage\\ (MB)\end{tabular}} &
  \multicolumn{1}{c|}{\begin{tabular}[c]{@{}c@{}}GPU RAM Usage\\ (MB)\end{tabular}} &
  \multicolumn{1}{c|}{\begin{tabular}[c]{@{}c@{}}Power Consumption\\ (W)\end{tabular}} &
  \multicolumn{1}{c|}{AUC-ROC} &
  \multicolumn{1}{c}{\begin{tabular}[c]{@{}c@{}}Inference Frequency \\ (Hz)\end{tabular}} \\ 
\hline \hline
\multirow{7}{*}{\begin{tabular}[c]{@{}c@{}}Jetson \\ Xavier\\ NX\end{tabular}} 
& Idle                  & 36.465 & 52.100 & 5,130.219 & 537.235 & 5.851  & . & .      \\ \cline{2-9} 
& \gls{arlstm}          & 62.311 & 97.700 & 5,669.830 & 872.374 & 11.288  & 0.719 & 5.200  \\ \cline{2-9} 
& \gls{gbrf}            & 61.499 & 53.000 & 5,518.050 & 528.416 & 6.108  & 0.655 & 20.575  \\ \cline{2-9} 
& \gls{ae}              & 53.023 & 79.400 & 5,276.139 & 807.528 & 6.010  & 0.810 & 2.247  \\ \cline{2-9} 
& \gls{knn}             & 92.547 & 55.700 & 5,076.605 & 526.844 & 7.208  & 0.718 & 1.116  \\ \cline{2-9} 
& Isolation Forest      & 51.122 & 64.700 & 4,859.356 & 526.673 & 5.777  & 0.629 & 4.568  \\ \cline{2-9} 
& \cellcolor{lightgray} \toolname & \cellcolor{lightgray} 52.420 & \cellcolor{lightgray} 70.600 & \cellcolor{lightgray} 5,488.874 & \cellcolor{lightgray} 1,005.369 & \cellcolor{lightgray} 6.333  & \cellcolor{lightgray} 0.844 & \cellcolor{lightgray} 14.937  \\ 
\hline \hline
\multirow{7}{*}{\begin{tabular}[c]{@{}c@{}}Jetson \\ AGX \\ Orin\end{tabular}} 
& Idle                  & 4.875  & 0.000  & 3,916.715 & 243.289 & 7.522  & .     & .      \\  \cline{2-9} 
& \gls{arlstm}          & 10.744 & 87.200 & 4,741.666 & 761.107 & 11.139 & 0.719 & 8.687  \\ \cline{2-9} 
& \gls{gbrf}            & 10.475 & 15.900 & 4,279.286 & 245.287 & 9.741  & 0.655 & 44.128 \\ \cline{2-9} 
& \gls{ae}              & 10.548 & 51.800 & 4,882.850 & 699.010 & 10.168 & 0.810 & 4.284  \\\cline{2-9} 
& \gls{knn}             & 91.506 & 0.000  & 4,201.195 & 243.289 & 16.887 & 0.718 & 4.754  \\ \cline{2-9} 
& Isolation Forest      & 10.648 & 0.000  & 3,990.171 & 243.289 & 9.169  & 0.629 & 10.732  \\ 

\cline{2-9} 
& \cellcolor{lightgray} \toolname & \cellcolor{lightgray} 10.399 & \cellcolor{lightgray} 70.100 & \cellcolor{lightgray} 5,167.490 & \cellcolor{lightgray} 954.701 & \cellcolor{lightgray} 10.220 & \cellcolor{lightgray} 0.844 & \cellcolor{lightgray} 26.461 \\ 
\hline
\end{tabular}

    }
   \vspace{-.5em}
\end{table*}

\begin{figure}[t]
    \centering
    \vspace{-.5em}
    \includegraphics[width=0.9\columnwidth]{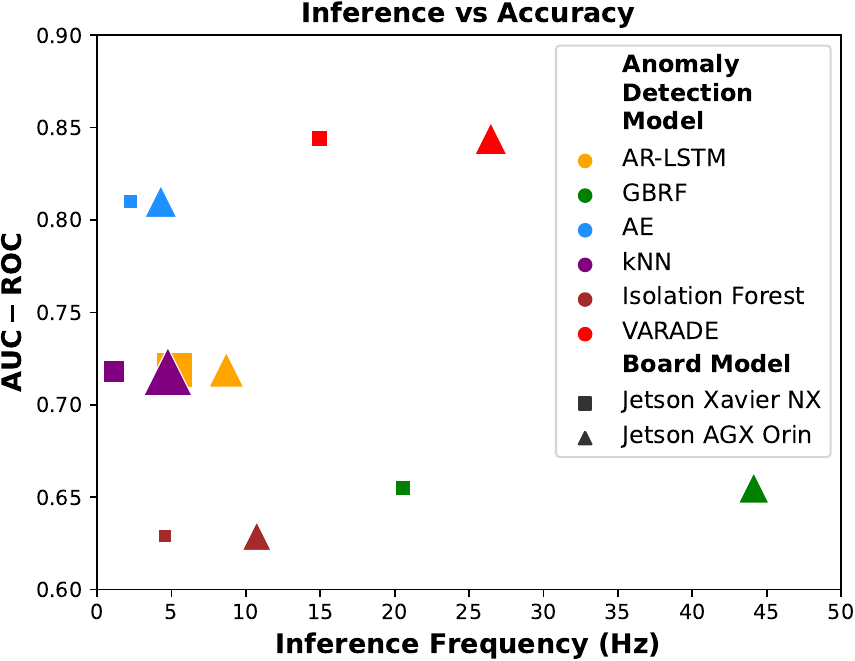}
    \vspace{-.5em}
    \caption{Inference Frequency vs Accuracy of the anomaly detection models (identified by the marker color). Marker shape represents the adopted edge device (square for Jetson Xavier NX, and triangle for Jetson AGX Orin), while marker size is proportional to power consumption.}
    \label{fig:inference-x-accuracy}
    \vspace{-1em}
\end{figure}

\vspace{-0.5em}
\paragraph{\textbf{Jetson Xavier NX}}
\toolname{} placed first in terms of accuracy, with an AUC-ROC score of 0.84, with an improvement of 4\% \wrt{} the second most performing model (\gls{ae}) and of 13\% \wrt{} the 
third most performing one (\gls{arlstm}).
Interestingly, these improvements correspond also to a higher inference frequency, improved 7 times \wrt{} \gls{ae} and 3 times
\wrt{} \gls{arlstm}.

Considering inference frequency, \toolname{} placed second, with 15 Hz against 
20 Hz obtained by \gls{gbrf}. However, when considering the AUC-ROC scores, \toolname{} offers
an improvement of almost 20\% \wrt\ \gls{gbrf}. 
Looking at RAM usage, we note that all the models use almost the same amount of memory, while \toolname{} uses a higher amount of GPU RAM (500 MB). This is not a limitation for the applicability of \toolname{}, as the total amount of memory used is under 40\%, thus leaving enough space for larger
anomaly models or other applications. 
Another important parameter for an edge device is the power required to operate, as the device could operate in conditions with limited power. 
Almost all the models have comparable performance in terms of power consumption, 
except \gls{arlstm} (for its high usage of the GPU), and \gls{knn} (for its high usage of CPU).

\paragraph{\textbf{Jetson AGX Orin}}
Analyzing the results obtained on the Jetson AGX Orin, we can note that the
results are similar to the ones obtained with the Jetson
Xavier
NX but with a different scale. 
We can see that inference frequency is more or less doubled for all the 
anomaly detection models, but 
the overall ranking remains the same, with \gls{gbrf} in the first position and 
\toolname{} in the second position.
A significant difference relies in the GPU usage, as in this case the TensorFlow planner decided to run \gls{knn} and Isolation Forest on the CPU due to
the higher number of CPU cores.

\cref{fig:inference-x-accuracy} highlights the characteristics of the different configurations by depicting the ratio between the Inference Frequency 
and the AUC-ROC scores of the tested anomaly detection models. 

Looking at the results obtained on the two boards, we can draw the following conclusions.
The two anomaly detection models less suitable (in our case study) for anomaly detection on the edge are the \gls{knn} and the \gls{arlstm}.
\gls{knn} is an algorithm that cannot fully benefit from GPU parallelism (especially with a few channels, as in our case study). On one side, this problem can be solved by exploiting CPU parallelism, but on the other side, edge devices have limited computation power, leading to high power draw and limited CPU available to run other jobs.
\gls{arlstm} is based on a memory-intensive architecture that is not designed to work in a constrained environment with high throughput requirements. 
In fact, with both the boards, we can note a high GPU usage, which could seem a positive factor but leads in fact to low inference speed. 

At the same time, we can note that \toolname{} (in red) shows the best accuracy without sacrificing too 
much performance on the inference speed, thus offering the best trade-off on both edge devices.
This demonstrates its applicability on constrained devices such as the Jetson 
Xavier NX, as it offers a significant improvement in accuracy with a minimal loss in the inference speed, still meeting the resource constraints imposed by an edge device.

\section{Conclusions and future works}
\label{sec:conclusions}
\balance
In this research we introduced \toolname{}, a variational based autoregressive system, to address the challenges posed by real-time anomaly detection on the edge. When benchmarked against conventional algorithms, \toolname{} demonstrated superior performance, while maintaining a significantly higher inference speed compared to other anomaly detection techniques. This positions VARADE as a promising solution, especially in applications that can benefit from the ability to detect complex anomalies. Future works will include experimenting with a larger set of different use cases, to stress the flexibility of our method.
Thus, we plan to integrate \toolname{} within the manufacturing control loop, enabling preventive anomaly detection to activate high-level reconfiguration strategies.

\bibliographystyle{ACM-Reference-Format}
\bibliography{literature}

\end{document}